\documentclass[11pt,a4paper]{article}
\usepackage[hyperref]{emnlp-ijcnlp-2019}
\usepackage{times}
\usepackage{latexsym}
\usepackage{graphicx}
\usepackage{url}
\usepackage{multirow}
\usepackage{booktabs}
\usepackage{amsmath}
\usepackage{tabularx}
\usepackage{ragged2e}
\usepackage{footnote}
\aclfinalcopy 

\title{Addressing Semantic Drift in Question Generation \\ for Semi-Supervised Question Answering}

\author{Shiyue Zhang $\;\;\;\;$ Mohit Bansal \\
  UNC Chapel Hill  \\
  {\tt \{shiyue, mbansal\}@cs.unc.edu} 
}

\date{}

\begin{document}
\maketitle
\begin{abstract}
 Text-based Question Generation (QG) aims at generating natural and relevant questions that can be answered by a given answer in some context. Existing QG models suffer from a ``semantic drift'' problem, i.e., the semantics of the model-generated question drifts away from the given context and answer. In this paper, we first propose two semantics-enhanced rewards obtained from downstream question paraphrasing and question answering tasks to regularize the QG model to generate semantically valid questions. Second, since the traditional evaluation metrics (e.g., BLEU) often fall short in evaluating the quality of generated questions, we propose a QA-based evaluation method which measures the QG model's ability to mimic human annotators in generating QA training data. Experiments show that our method achieves the new state-of-the-art performance w.r.t. traditional metrics, and also performs best on our QA-based evaluation metrics. Further, we investigate how to use our QG model to augment QA datasets and enable semi-supervised QA. We propose two ways to generate synthetic QA pairs: generate new questions from existing articles or collect QA pairs from new articles. 
 We also propose two empirically effective strategies, a data filter and mixing mini-batch training, to properly use the QG-generated data for QA.
 Experiments show that our method improves over both BiDAF and BERT QA baselines, even without introducing new articles.\footnote{Code and models publicly available at: \url{https://github.com/ZhangShiyue/QGforQA}}
\end{abstract}

\section{Introduction}
In contrast to the rapid progress shown in Question Answering (QA) tasks \cite{rajpurkar2016squad, joshi2017triviaqa, yang2018hotpotqa}, the task of Question Generation (QG) remains understudied and challenging. However, as an important dual task to QA, QG can not only be used to augment QA datasets \cite{duan-etal-2017-question}, but can also be applied in conversation and education systems \cite{heilman2010good, lindberg2013generating}. Furthermore, given that existing QA models often fall short by doing simple word/phrase matching rather than true comprehension \cite{jia2017adversarial}, the task of QG, which usually needs complicated semantic reasoning and syntactic variation, should be another way to encourage true machine comprehension \cite{lewis2018generative}. Recently, we have seen an increasing interest in the QG area, with mainly three categories: Text-based QG \cite{du2017learning, zhao2018paragraph}, Knowledge-Base-based QG \cite{reddy2017generating, serban2016generating}, and Image-based QG \cite{li2018visual, jain2017creativity}. Our work focuses on the Text-based QG branch. 

\begin{figure}[t!]
\small
\begin{tabularx}{0.46\textwidth}{X}
\toprule
\textbf{Context}: ...during the age of enlightenment, philosophers such as \underline{\textbf{john locke}} advocated the principle in their writings, whereas others, such as thomas hobbes, strongly opposed it. montesquieu was one of the foremost supporters of separating the legislature, the executive, and the judiciary... \\
\midrule
\textbf{Gt}: who was an advocate of separation of powers? \\
\textbf{Base}: who opposed the principle of enlightenment? \\
\textbf{Ours}: who advocated the principle in the age of enlightenment? \\
\bottomrule
\end{tabularx}
\vspace{-7pt}
\caption{An examples of the ``semantic drift'' issue in Question Generation (``Gt'' is short for ``ground truth''). }
\vspace{-20pt}
\label{tab:sem-drift}
\end{figure}

Current QG systems follow an attention-based sequence-to-sequence structure, taking the paragraph-level context and answer as inputs and outputting the question. However, we observed that these QG models often generate questions that semantically drift away from the given context and answer; we call this the ``semantic drift'' problem. As shown in Figure~\ref{tab:sem-drift}, the baseline QG model generates a question that has almost contrary semantics with the ground-truth question, and the generated phrase ``the principle of enlightenment'' does not make sense given the context. We conjecture that the reason for this ``semantic drift'' problem is because the QG model is trained via teacher forcing only, without any high-level semantic regularization. Hence, the learned model behaves more like a question language model with some loose context constraint, while it is unaware of the strong requirements that it should be closely grounded by the context and should be answered by the given answer. Therefore, we propose two semantics-enhanced rewards to address this drift: \textbf{QPP} and \textbf{QAP}. Here, \textbf{QPP} refers to \textbf{Q}uestion \textbf{P}araphrasing \textbf{P}robability, which is the probability of the generated question and the ground-truth question being paraphrases; \textbf{QAP} refers to \textbf{Q}uestion \textbf{A}nswering \textbf{P}robability, which is the probability that the generated question can be correctly answered by the given answer. We regularize the generation with these two rewards via reinforcement learning. Experiments show that these two rewards can significantly improve the question generation quality separately or jointly, and achieve the new state-of-the-art performance on the SQuAD QG task.

Next, in terms of QG evaluation, previous works have mostly adopted popular automatic evaluation metrics, like BLEU, METEOR, etc. However, we observe that these metrics often fall short in properly evaluating the quality of generated questions. First, they are not always correlated to human judgment about answerability \cite{nema2018towards}. Second, since multiple questions are valid but only one reference exists in the dataset, these traditional metrics fail to appropriately score question paraphrases and novel generation (shown in Figure~\ref{tab-example}).  Therefore, we introduce a QA-based evaluation method that directly measures the QG model's ability to mimic human annotators in generating QA training data, because ideally, we hope that the QG model can act like a human to ask questions. We compare different QG systems using this evaluation method, which shows that our semantics-reinforced QG model performs best. However, this improvement is relatively minor compared to our improvement on other QG metrics, which indicates improvement on typical QG metrics does not always lead to better question annotation by QG models for generating QA training set.

Further, we investigate how to use our best QG system to enrich QA datasets and perform semi-supervised QA on SQuADv1.1 \cite{rajpurkar2016squad}. Following the back-translation strategy that has been shown to be effective in Machine Translation \cite{sennrich2016improving} and Natural Language Navigation \cite{fried2018speaker, tan2019learning}, we propose two methods to collect synthetic data. First, since multiple questions can be asked for one answer while there is only one human-labeled ground-truth, we make our QG model generate new questions for existing context-answer pairs in SQuAD training set, so as to enrich it with paraphrased and other novel but valid questions. Second, we use our QG model to label new context-answer pairs from new Wikipedia articles. However, directly mixing synthetic QA pairs with ground-truth data will not lead to improvement. Hence, we introduce two empirically effective strategies: one is a ``data filter'' based on QAP (same as the QAP reward) to filter out examples that have low probabilities to be correctly answered; the other is a ``mixing mini-batch training'' strategy that always regularizes the training signal with the ground-truth data. Experiments show that our method improves both BiDAF \cite{seo2016bidirectional, clark2018simple} and BERT \cite{devlin2018bert} QA baselines by 1.69/1.27 and 1.19/0.56 absolute points on EM/F1, respectively; even without introducing new articles, it can bring 1.51/1.13 and 0.95/0.13 absolute improvement, respectively.

\section{Related Works}
\paragraph{Question Generation} Early QG studies focused on using rule-based methods to transform statements to questions \cite{heilman2010good, lindberg2013generating, labutov2015deep}. Recent works adopted the attention-based sequence-to-sequence neural model \cite{bahdanau2014neural} for QG tasks, taking answer sentence as input and outputting the question \cite{du2017learning, zhou2017neural}, which proved to be better than rule-based methods. Since human-labeled questions are often relevant to a longer context, later works leveraged information from the whole paragraph for QG, either by extracting additional information from the paragraph \cite{du2018harvesting, song2018leveraging, liu2019learning} or by directly taking the whole paragraph as input \cite{zhao2018paragraph, kim2018improving, sun2018answer}. A very recent concurrent work applied the large-scale language model pre-training strategy for QG and also achieved a new state-of-the-art performance \cite{dong2019unified}. However, the above models were trained with teacher forcing only. To address the exposure bias problem, some works applied reinforcement learning taking evaluation metrics (e.g., BLEU) as rewards \cite{song2017unified, kumar2018framework}.  \citet{yuan2017machine} proposed to use a language model's perplexity ($R_{PPL}$) and a QA model's accuracy ($R_{QA}$) as two rewards but failed to get significant improvement. Their second reward is similar to our QAP reward except that we use QA probability rather than accuracy as the probability distribution is more smooth. \citet{hosking2019evaluating} compared a set of different rewards, including $R_{PPL}$ and $R_{QA}$, and claimed none of them improved the quality of generated questions. For QG evaluation, even though some previous works conducted human evaluations, most of them still relied on traditional metrics (e.g., BLEU). However, \citet{nema2018towards} pointed out the existing metrics do not correlate with human judgment about answerability, so they proposed ``Q-metrics'' that mixed traditional metrics with an ``answerability'' score. In our work, we will show QG results on traditional metrics, Q-metrics, as well as human evaluation, and also propose a QA-based QG evaluation.

\paragraph{Question Generation for QA} As the dual task of QA, QG has been often proposed for improving QA. Some works have directly used QG in QA models' pipeline \cite{duan-etal-2017-question, dong2017learning, lewis2018generative}. Some other works enabled semi-supervised QA with the help of QG. \citet{tang2017question} applied the ``dual learning'' algorithm \cite{he2016dual} to learn QA and QG jointly with unlabeled texts.  \citet{yang2017semi} and \citet{tang2018learning} followed the GAN \cite{goodfellow2014generative} paradigm, taking QG as a generator and QA as a discriminator, to utilize unlabeled data. \citet{sachan2018self} proposed a self-training cycle between QA and QG. However, these works either reduced the ground-truth data size or simplified the span-prediction QA task to answer sentence selection. \citet{dhingra2018simple} collected 3.2M cloze-style QA pairs to pre-train a QA model, then fine-tune with the full ground-truth data which improved a BiDAF-QA baseline. In our paper, we follow the back-translation \cite{sennrich2016improving} strategy to generate new QA pairs by our best QG model to augment SQuAD training set. Further, we introduce a data filter to remove poorly generated examples and a mixing mini-batch training strategy to more effectively use the synthetic data. Similar methods have also been applied in some very recent concurrent works~\cite{dong2019unified, alberti2019synthetic} on SQuADv2.0. The main difference is that we also propose to generate new questions from existing articles without introducing new articles.

\section{Question Generation}
\subsection{Base Model}
\begin{figure}
    \centering
    \includegraphics[width=0.4\textwidth]{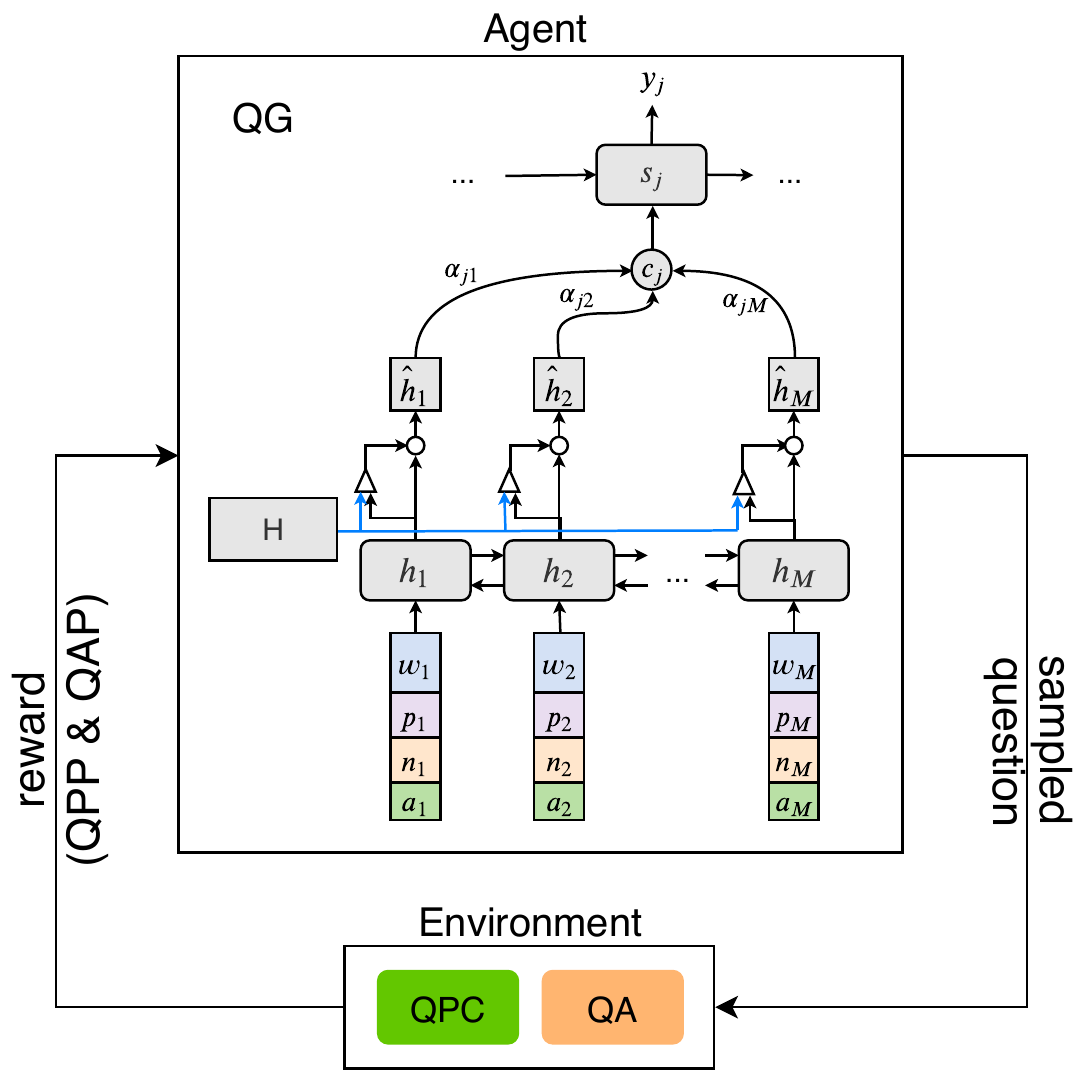}
    \vspace{-7pt}
    \caption{The architecture of our semantics-reinforced QG model.}
    \vspace{-10pt}
    \label{fig:qg}
\end{figure}

We first introduce our base model which mainly adopts the model architecture from the previous state-of-the-art \cite{zhao2018paragraph}. The differences are that we introduce two linguistic features (POS \& NER), apply deep contextualized word vectors, and tie the output projection matrix with the word embedding matrix. Experiments showed that with these additions, our base model results surpass the results reported in \citet{zhao2018paragraph} with significant margins. Our base model architecture is shown in the upper box in Figure~\ref{fig:qg} and described as follow. If we have a paragraph $p=\{x_i\}_{i=1}^M$ and an answer $a$ which is a sub-span of $p$, the target of the QG task is to generate a question $q=\{y_j\}_{j=1}^N$ that can be answered by $a$ based on the information in $p$. 

\paragraph{Embedding}  The model first concatenates four word representations: word vector, answer tag embedding, Part-of-Speech (POS) tag embedding, and Name Entity (NER) tag embedding, i.e., $e_i=[w_i, a_i, p_i, n_i]$. For word vectors, we use the deep contextualized word vectors from ELMo \cite{peters2018deep} or BERT \cite{devlin2018bert}. 
The answer tag follows the BIO\footnote{``B'', for ``Begin'', tags the start token of the answer span; ``I'', for ``Inside'', tags other tokens in the answer span; ``O'', for ``Other'', tags other tokens in the paragraph.} tagging scheme.

\paragraph{Encoder} The output of the embedding layer is then encoded by a two-layer bi-directional LSTM-RNNs, resulting in a list of hidden representations $H$. At any time step $i$, the representation $h_i$ is the concatenation of $\overrightarrow{h_i}$ and $\overleftarrow{h_i}$.
\begin{equation}
\begin{gathered}
   \overrightarrow{h}_i=\overrightarrow{LSTM}([e_i; \overrightarrow{h}_{i-1}] ) \\
   \overleftarrow{h}_i=\overleftarrow{LSTM}([e_i; \overleftarrow{h}_{i+1}] ) \\
    H = [\overrightarrow{h_i}, \overleftarrow{h_i}]_{i=1}^M
\end{gathered}
\end{equation}

\paragraph{Self-attention} A gated self-attention mechanism \cite{wang2017gated} is applied to $H$ to aggregate the long-term dependency within the paragraph. $\alpha_i$ is an attention vector between $h_i$ and each element in $H$; $u_i$ is the self-attention context vector for $h_i$; $h_i$ is then updated to $f_i$ using $u_i$; a soft gate $g_i$ decides how much the update is applied. $\hat{H}=[\hat{h}_i]_{i=1}^M$ is the output of this layer.
\begin{equation}
\begin{gathered}
    u_i = H\alpha_i, \alpha_i = softmax(H^TW^uh_i) \\
    f_i = tanh(W^f[h_i; u_i]) \\
    g_i = sigmoid(W^g[h_i; u_i]) \\
    \hat{h}_i = g_i*f_i + (1-g_i)*h_i
\end{gathered}
\end{equation}

\paragraph{Decoder} The decoder is another two-layer uni-directional LSTM-RNN. An attention mechanism dynamically aggregates $\hat{H}$ at each decoding step to a context vector $c_j$ which is then used to update the decoder state $s_j$.
\begin{equation}
\begin{gathered}
    c_j = \hat{H}\alpha_j, \alpha_j = softmax(\hat{H}^TW^as_j) \\
    \tilde{s}_j = tanh(W^c[c_j; s_j]) \\
    s_{j+1}=LSTM([y_j; \tilde{s}_{j}] )
\end{gathered}
\end{equation}
The probability of the target word $y_j$ is computed by a maxout neural network.
\begin{equation}
\begin{gathered}
    \tilde{o}_j = tanh (W^o[c_j; s_j]) \\
    o_j = [max \{\tilde{o}_{j, 2k-1}, \tilde{o}_{j, 2k}\}]_k \\
    p(y_j|y_{<j}) = softmax(W^eo_j)
\end{gathered}
\end{equation}
In practice, we keep the weight matrix $W^e$ the same as the word embedding matrix and fix it during training. Furthermore, we apply a ``pointer network'' \cite{gu2016incorporating} to enable the model to copy words from input.

\subsection{Semantics-Reinforced Model}
To address the ``semantic drift" problem shown in Figure~\ref{tab:sem-drift}, we propose two semantics-enhanced rewards to regularize the generation to focus on generating semantically valid questions. 

\paragraph{QPP Reward} To deal with the ``exposure bias'' problem, many previous works directly used the final evaluation metrics (e.g., BLEU) as rewards to train the generation models \cite{rennie2017self, paulus2017deep}. However, these metrics sometimes fail to evaluate equally to question paraphrases and thus provide inaccurate rewards. Hence, we propose to use a pre-trained question paraphrasing classification (QPC) model to provide paraphrasing probability as a reward. Since paraphrasing is more about semantic similarity than superficial word/phrase matching, it treats question paraphrases more fairly (Example 1 in Figure~\ref{tab-example}).
Therefore, we first train a QPC model with Quora Question Pairs dataset. Next, we take it as an environment, and the QG model will interact with it during training to get the probability of the generated question and the ground-truth question being paraphrases as the reward. 

\paragraph{QAP Reward} Two observations motivate us to introduce QAP reward. First, in a paragraph, usually, there are several facts relating to the answer and can be used to ask questions. Neither the teacher forcing or the QPP reward can favor this kind of novel generation (Example 2 in Figure~\ref{tab-example}). Second, we find semantically-drifted questions are usually unanswerable by the given answer. Therefore, inspired by the dual learning algorithm \cite{he2016dual}, we propose to take the probability that a pre-trained QA model can correctly answer the generated question as a reward, i.e., $p(a^*|q^s; p)$, where $a^*$ is the ground-truth answer and $q^s$ is a sampled question. Using this reward, the model can not only gets positive rewards for novel generation but also be regularized by the answerability requirement. Note that, this reward is supposed to be carefully used because the QG model can cheat by greedily copying words in/near the answer to the generated question. In this case, even though high QAPs are achieved, the model loses the question generation ability. 
\begin{figure*}[t!]
\begin{center}
\small
\begin{tabularx}{\textwidth}{Xcccc}
\toprule Example 1: Fail to score equally to paraphrases  & BLEU4 & Q-BLEU1 & QPP & QAP  \\ 
\midrule 
Context: ...the university first offered graduate degrees , in the form of a master of arts ( ma ) , in the the  \underline{\textbf{1854}} -- 1855 academic year ...  & & & & \\
Gt: in what year was a master of arts course first offered ? & & & & \\
Gen1: in what year did the university first offer a master of arts ? & 37.30 & 79.39& 49.71& 34.09\\
Gen2: when did the university begin offering a master of arts ? & 29.58& 47.50 & 46.12& 18.18\\
\midrule
Example 2: Fail to score appropriately to novel generation & & & & \\
\midrule 
Context: ...in \underline{\textbf{1987}} , when some students believed that the observer began to show a conservative bias , a liberal newspaper , common sense was published... & & & & \\
Gt: in what year did the student paper common sense begin publication ? & & & & \\
Gen1: in what year did common sense begin publication ? & 56.29 & 85.77 & 92.28& 93.94\\
Gen2: when did the observer begin to show a conservative bias ? & 15.03 & 21.11 & 13.44& 77.15\\
\toprule
\end{tabularx}
\end{center}
\vspace{-15pt}
\caption{\label{font-table} Two examples of where QPP and QAP improve in question quality evaluation. }
\vspace{-10pt}
\label{tab-example}
\end{figure*}
\paragraph{Policy Gradient}
To apply these two rewards, we use the REINFORCE algorithm \cite{williams1992simple}  to learn a generation policy $p_\theta$ defined by the QG model parameters $\theta$. We minimize the loss function $L_{RL} = -E_{q^s \sim p_\theta}[r(q^s)]$, where $q^s$ is a sampled question from the model's output distribution. Due to the non-differentiable sampling procedure, the gradient is approximated using a single sample with some variance reduction baseline $b$:
\begin{equation}
    \bigtriangledown_\theta L_{RL} = -(r(q^s) - b)\bigtriangledown_\theta log p_\theta(q^s)
\end{equation}
We follow the effective SCST strategy \cite{rennie2017self} to take the reward of greedy search result $q^g$ as the baseline, i.e., $b = r(q^g)$. However, only using this objective to train QG will result in poor readability, so we follow the mixed loss setting \cite{paulus2017deep}: $L_{mixed} = \gamma L_{RL} + (1-\gamma)L_{ML}$. In practice, we find the mixing ratio $\gamma$ for QAP reward should be lower, i.e., it needs more regularization from teacher forcing, so that it can avoid the undesirable cheating issue mentioned above. Furthermore, we also apply the multi-reward optimization strategy \cite{pasunuru2018multi} to train the model with two mixed losses alternately with an alternate rate $n:m$, i.e., train with $L_{mixed}^{qpp}$ for $n$ mini-batches, then train with $L_{mixed}^{qap}$ for $m$ mini-batches, repeat until convergence. $n$ and $m$ are two hyper-parameters. 
\begin{equation}
\begin{gathered}
    L_{mixed}^{qpp} = \gamma^{qpp} L_{RL}^{qpp} + (1-\gamma^{qpp})L_{ML} \\
    L_{mixed}^{qap} = \gamma^{qap} L_{RL}^{qap} + (1-\gamma^{qap})L_{ML}
\end{gathered}
\end{equation}
Experiments show that these two rewards can significantly improve the QG performance separately or jointly, and we achieve new state-of-the-art QG performances, see details in Section~\ref{sec:res}.

\subsection{QA-Based QG Evaluation}
\label{sec:qa-eval}
Inspired by the idea that ``a perfect QG model can replace humans to ask questions'', we introduce a QA-based evaluation method that measures the quality of a QG model by its ability to mimic human annotators in labeling training data for QA models. The evaluation procedure is described as follows. First, we sample some unlabeled Wikipedia paragraphs with pre-extracted answer spans from HarvestingQA dataset \cite{du2018harvesting}. Second, we make a QG model act as an ``annotator" to annotate a question for each answer span. Third, we train a QA model using this synthetic QA dataset. Lastly, we use the QA model's performance on the original SQuAD development set as the evaluation for this QG model. The higher this QA performance is, the better the QG model mimics a human's question-asking ability. 
We believe that this method provides a new angle to evaluate QG model's quality and also a more reliable way to choose QG models to conduct data augmentation and semi-supervised QA.

\section{Semi-Supervised Question Answering}

\begin{figure*}
    \centering
    \includegraphics[width=0.75\textwidth]{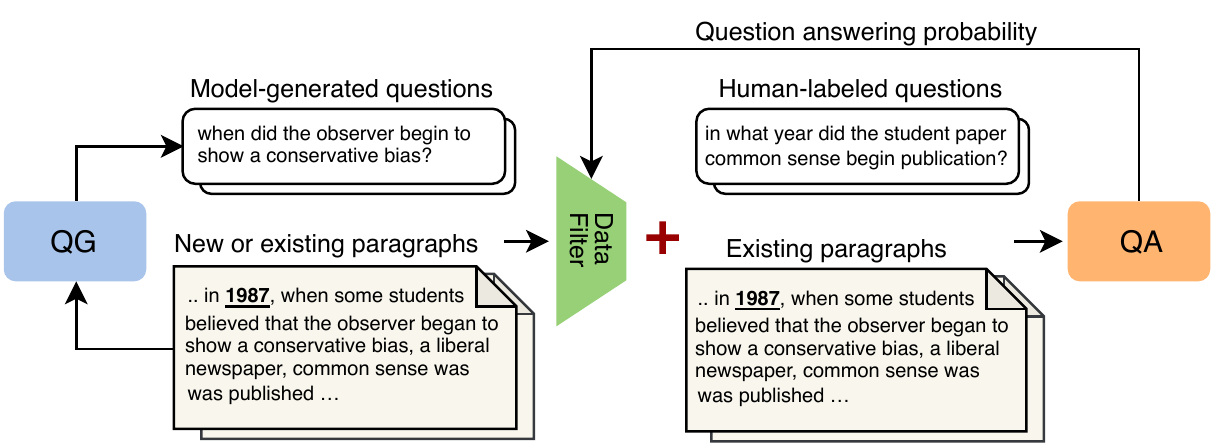}
    \vspace{-5pt}
    \caption{Semi-supervised QA: First, a trained QG model generates questions from new or existing paragraphs building up a synthetic QA dataset; Second, a data filter filters out low-QAP synthetic examples and augment the rest to human-labeled QA pairs; Lastly, the QA model is trained with the enlarged QA dataset. \vspace{-9pt}}
    \label{fig:semi-qa}
\end{figure*}

Since one of the major goals of developing QG systems is to generate new QA pairs and augment QA datasets, we investigate how to use our QG system to act as a question annotator, collect new QA pairs, and conduct semi-supervised QA.
Figure~\ref{fig:semi-qa} illustrates the overall procedure of our semi-supervised QA approach. 

\subsection{Synthetic Data Generation}
To generate synthetic QA pairs, we follow the effective ``back translation'' approach proposed in Neural Machine Translation (NMT) \cite{sennrich2016improving}. In NMT, the back translation method first obtains synthetic source sentences by running a pre-trained target-to-source translation model on a monolingual dataset of the target language; then, it combines the synthetic and ground-truth translation pairs to train the desired source-to-target translation model. Similarly, in the QA scenario, the paragraph/answer can be viewed as the ``target sentence'', while the question can be taken as the ``source sentence''. One tricky difference is that even if the paragraphs can be easily obtained from Wikipedia, there are no answer span labels. Therefore, we use two sources to generate questions from, as discussed below.  

\paragraph{Generate from Existing Articles} In SQuAD \cite{rajpurkar2016squad}, each context-answer pair only has one ground-truth question. However, usually, multiple questions can be asked. The diversity lies in question paraphrasing and different facts in the context that can be used to ask the question. Therefore, without introducing new Wikipedia articles, we make our QG model generate diverse questions for the existing context-answer pairs in SQuAD training set by keeping the all beam search outputs for each example. 

\paragraph{Generate from New Articles} To use unlabeled Wikipedia articles for data augmentation, an automatic answer extractor is indispensable. Some previous works have proposed methods to detect key phrases from a paragraph and automatically extract potential answer spans \cite{yang2017semi, du2018harvesting, subramanian2018neural}. Instead of building up our answer extractor, we directly take advantage of the released HarvestingQA dataset. It contains 1.2M synthetic QA pairs, in which both the answer extractor and the QG model were proposed by \citet{du2018harvesting}. We use their paragraphs with answer span labels but generate questions with our QG models, and only use their questions for comparison.

\subsection{Synthetic Data Usage}
In practice, we find that directly mixing the synthetic data with the ground-truth data does not improve QA performance. We conjecture the reason is that some poor-quality synthetic examples mislead the learning process of the QA model. Therefore, we propose two empirical strategies to better utilize synthetic data.
\paragraph{QAP Data Filter} In ``self-training'' literature, similar issues have been discussed that using model-labeled examples to train the model will amplify the model's error. Later works proposed co-training or tri-training that uses two or three models as judges and only keeps examples that all models agree on \cite{blum1998combining, zhou2005tri}. \citet{sachan2018self} also designed question selection oracles based on curriculum learning strategy in their QA-QG self-training circle. In this paper, we simply design a data filter based on our QAP measure (same definition as the QAP reward) to filter poor-quality examples. We think if one question-answer pair has a low QAP, i.e., the probability of the answer given the question is low, it is likely to be a mismatched pair. Hence, we filter synthetic examples with $QAP < \epsilon$, where $\epsilon$ is a hyper-parameter that we will tune for different synthetic datasets.

\begin{savenotes}
\begin{table*}[t!]
\begin{center}
\small
\begin{tabular}{lllllll}
\toprule  & BLEU4 & METEOR & ROUGE-L & Q-BLEU1 & QPP & QAP  \\ \hline
\citet{du2018harvesting} & 15.16 & 19.12 & -- & -- & -- & -- \\
\citet{zhao2018paragraph}\footnote{They actually used the reversed dev-test setup as opposed to the original setup used in \citet{du2017learning} and \citet{du2018harvesting} (see Section 3.1 in \citet{zhao2018paragraph}). Thus, we also conducted the reversed dev-test setup and our QPP\&QAP model yields BLEU4/METEOR/ROUGE-L=20.76/24.20/48.91.} & 16.38 & 20.25 & 44.48 & -- & -- & -- \\
\midrule
Our baseline (w. ELMo) & 17.00 & 21.44 & 45.89 & 47.80 & 27.29 & 45.15 \\
 + BLEU4 & 17.72 & 22.13 & 46.52 & 49.07 & 27.09 & 45.96  \\
 + METEOR & 17.84 & 22.41 & 46.18 & 49.09 & 26.70 & 46.52 \\
 + ROUGE-L & 17.78 & 22.28 & 46.51 & 49.23 & 27.06 & 46.31  \\
\midrule
 + QPP & 18.25 & 22.62 & 46.45 & 49.59 & \bf 28.13 & 47.63  \\
 + QAP & 18.12 & 22.52 & 46.45 & 49.27 & 27.49 & \bf 48.76  \\
 + QPP\&QAP & \bf 18.37 & \bf 22.65 & \bf 46.68 & \bf 49.63 & 28.03 & 48.37 \\
\bottomrule
\end{tabular}
\end{center}
\vspace{-12pt}
\caption{The performance of different QG models.} 
\vspace{-5pt}
\label{tab-qg-res}
\end{table*}
\end{savenotes}

\paragraph{Mixing Mini-Batch Training} When conducting semi-supervised learning, we do not want gradients from ground-truth data are overwhelmed by synthetic data. Previous works \cite{fried2018speaker, dhingra2018simple} proposed to first pre-train the model with synthetic data and then fine-tune it with ground-truth data. However, we find when the synthetic data size is small (e.g., similar size as the ground-truth data), catastrophic forgetting will happen during fine-tuning, leading to similar results as using ground-truth data only. Thus, we propose a ``mixing mini-batch'' training strategy, where for each mini-batch we combine half mini-batch ground-truth data with half mini-batch synthetic data, which keeps the data mixing ratio to 1:1 regardless of what the true data size ratio is. In this way, we can have the training process generalizable to different amounts of synthetic data and keep the gradients to be regularized by ground-truth data.

\section{Experiment Setup}
\paragraph{Datasets} For QG, we use the most commonly used SQuAD QG dataset first used by \citet{du2017learning}. For QA-based QG evaluation, we obtain unlabeled paragraph and answer labels from HarvestingQA \cite{du2018harvesting}, and have different QG systems to label questions. For semi-supervised QA, we use SQuADv1.1 \cite{rajpurkar2016squad} as our base QA task, and split the original development set in half as our development and test set respectively. Plus, we make our QG model generate new questions from both SQuAD and HarvestingQA. We will sample 10\% -- 100\% examples from HarvestingQA which are denoted by H1-10 in our experiments.

\paragraph{Evaluation Metrics} For QG, we first adopt 3 traditional metrics (BLEU4/METEOR/ROUGE-L). Second, we apply the new Q-BLEU1 metric proposed by \citet{nema2018towards}. Moreover, we conduct a pairwise \textbf{human evaluation} between our baseline and QPP\&QAP model on MTurk. We gave the annotators a paragraph with an answer bold in context and two questions generated by two models (randomly shuffled). We asked them to decide which one is better or non-distinguishable. For both QA-based QG evaluation and semi-supervised QA, we follow the standard evaluation method for SQuAD to use Exact Match (EM) and F1. 

More details about datasets, evaluation metrics, human evaluation setup, and model implement details are provided in the Appendix.

\section{Results}
\label{sec:res}
\subsection{Question Generation}
\paragraph{Baselines} First, as shown in Table~\ref{tab-qg-res}, our baseline QG model obtains a non-trivial improvement over previous best QG system \cite{zhao2018paragraph} which proves the effectiveness of our newly introduced setups: introduce POS/NER features, use deep contexturalized word vectors (from ELMo or BERT), and tie output projection matrix with non-trainable word embedding matrix. Second, we apply three evaluation metrics as rewards to deal with the exposure bias issue and improve performance. All the metrics are significantly\footnote{The significance tests in this paper are conducted following the bootstrap test setup \cite{efron1994introduction}.} ($p<0.001$) improved except QPP, which supports that high traditional evaluation metrics do not always correlate to high semantic similarity.

\begin{table}
\begin{center}
\small
\begin{tabular}{lll}
\toprule QPP\&QAP & Our baseline & Tie  \\ 
\midrule
  160 & 131 & 9 \\
\toprule
\end{tabular}
\end{center}
\vspace{-12pt}
\caption{Pairwise human evaluation between our baseline and QPP\&QAP multi-reward model.}
\label{tab:human-eval}
\vspace{-1pt}
\end{table}

\begin{table}
\begin{center}
\small
\begin{tabular}{llll}
\toprule Data  & \citeauthor{du2018harvesting} & Our baseline & QPP \& QAP  \\ 
\midrule
H1 &  53.20/65.47 & 55.06/67.83 & \bf 55.89/68.26 \\
H2 &  53.40/66.28 & 56.23/\textbf{69.23} & \textbf{56.69}/69.19 \\
H3 &  53.12/65.57  & \textbf{57.14}/69.39 & 57.05/\textbf{70.17} \\
\midrule
S+H1 & 71.16/80.75  & 71.94/81.26 & \bf 72.20/81.44 \\
S+H2 & 72.02/81.00  & 72.03/81.38 & \bf 72.22/81.81 \\
S+H3 & 71.48/81.02  & 72.61/81.46 & \bf 72.69/82.22 \\
\toprule
\end{tabular}
\end{center}
\vspace{-12pt}
\caption{The QA-based evaluation results for different QG systems. The two numbers of each item in this table are the EM/F1 scores. All results are the performance on our QA test set. ``S'' is short for ``SQuAD''.}
\label{tab:qa-eval}
\vspace{-1pt}
\end{table}

\paragraph{Semantics-Reinforced Models} As shown in Table~\ref{tab-qg-res}, when using QAP and QPP separately, all metrics are significantly ($p<0.001$) improved over our baseline and all metrics except ROUGE-L are significantly ($p<0.05$) improved over the models using traditional metrics as rewards. After applying multi-reward optimization, our model performs
consistently best on BLEU4/METEOR/ROUGE-L and Q-BLEU1. Notably, using one of these two rewards will also improve the other one at the same time, but using both of them achieves a good balance between these two rewards without exploiting either of them and results in the consistently best performance on other metrics, which is a new state-of-the-art result. 
\textbf{Human Evaluation Results:} Table~\ref{tab:human-eval} shows the MTurk anonymous human evaluation study, where we do a pairwise comparison between our baseline and QPP\&QAP model. We collected 300 responses in total, 160 of which voted the QPP\&QAP model's generation is better, 131 of which favors the baseline model, and 9 of which selected non-distinguishable. 

\paragraph{QA-Based Evaluation} As shown in Table~\ref{tab:qa-eval}, we compare three QG systems using QA-based evaluation on three different amounts of synthetic data and their corresponding semi-supervised QA setups (without filter). It can be observed that both our baseline and our best QG model can significantly improve the synthetic data's QA performance, which means they can act as better ``annotators'' than the QG model proposed by \citet{du2018harvesting}. However, our best QG model only has a minor improvement over our baseline model, which means significant improvement over QG metrics does not guarantee significant better question annotation ability.  

\begin{savenotes}
\begin{table}[t!]
\begin{center}
\small
\begin{tabular}{lllll}
\toprule & Filter & Data Size\footnote{``Data Size'' counts the total number of examples in training set (after filter). In Table~\ref{tab:qa-comp}, ``New Data Size'' only counts \# examples generated from articles outside SQuAD.} &  EM & F1  \\ 
\midrule
 \multirow{4}{*}{\rotatebox{90} {H1 only}} & $\epsilon = 0.0$ & 120k & 54.55 & 67.91 \\
 & $\epsilon = 0.2$ & 84k & 
 61.18 & 71.65 \\
 & $\epsilon = 0.4$ & 69k & \bf 61.97 & \bf 72.48 \\
 & $\epsilon = 0.6$ & 55k & 60.38 & 70.51 \\
 & $\epsilon = 0.8$ & 40k & 57.47 & 66.48 \\
\midrule
 \multirow{4}{*}{\rotatebox{90} {SQuAD+H1}} & $\epsilon = 0.0$ & 207k & 72.97 & 82.18 \\
 & $\epsilon = 0.2$ & 171k & 73.88 & 82.72 \\
 & $\epsilon = 0.4$ & 156k & 73.47 & 82.62 \\
 & $\epsilon = 0.6$ & 142k & \bf 73.96 & \bf 82.81 \\
 & $\epsilon = 0.8$ & 127k & 73.65 & 82.77 \\
\toprule
\end{tabular}
\end{center}
\vspace{-12pt}
\caption{The effect of QAP-based synthetic data filter. We filter out the synthetic data with $QAP < \epsilon$. All results are the performance on our QA development set. } 
\label{tab:filter}
\vspace{-5pt}
\end{table}
\end{savenotes}

\subsection{Semi-Supervised Question Answering}
\paragraph{Effect of the data filter} As shown in Table~\ref{tab:filter}, when using synthetic data only, adding the data filter can significantly improve QA performance. In terms of semi-supervised QA, the improvement is relatively smaller, due to the regularization from ground-truth data, but still consistent and stable. 

\begin{table}[t!]
\begin{center}
\small
\begin{tabular}{lllll}
\toprule & Data & Data Size & EM & F1  \\ 
\midrule
\multirow{5}{*}{\rotatebox{90} {Dev set}} & SQuAD & 87k & 72.52 & 81.79  \\
& + Beam5  & 399k & 74.33 & 83.19 \\
& + Beam10  & 706k & 74.44 & \bf 83.23 \\
& + Beam15  & 853k & 74.25 & 82.75 \\
& + DivBeam10  & 595k & 74.44 & 83.30 \\
\midrule
\multirow{5}{*}{\rotatebox{90} {Dev set}} & + H1 & 142k  & 73.96 & 82.81 \\
& + H2 & 255k  & 74.19 & 82.84 \\
& + H4 & 424k  & 74.42 & 82.82 \\
& + H6 & 506k  & 74.27 & 82.97 \\
& + H8 & 705k  & \bf 74.64 & 83.14 \\
& + H10 & 930k  & 74.27 & 82.97 \\
\toprule
\multirow{4}{*}{\rotatebox{90} {Test set}} & SQuAD & 87k & 71.92 & 81.26  \\
& + Beam10 & 706k & 73.43 & 82.39 \\
& + H8 & 705k & \bf 73.61 & \bf 82.53 \\
& + Beam10 + H8 & 1.3M & 73.43  & 82.11 \\
\toprule
\end{tabular}
\end{center}
\vspace{-12pt}
\caption{The results of our semi-supervised QA method using a BiDAF-QA model.}
\label{tab:semi-res}
\vspace{-1pt}
\end{table}
 
\begin{table}[t!]
\begin{center}
\small
\begin{tabular}{llll}
\toprule Methods  & New Data Size & EM & F1  \\ 
\midrule
\citeauthor{dhingra2018simple} base & 0 & 71.54 & 80.69 \\
  +Cloze & 3.2M  & 71.86 & 80.80 \\
\midrule
Our base & 0  & 72.19 & 81.52 \\
 +Beam10 &  0  & 73.93 & 82.81 \\
 +H8 &  705k & 74.12 & 82.83 \\
\toprule
\end{tabular}
\end{center}
\vspace{-12pt}
\caption{The comparison with the previous semi-supervised QA method. All results are the performance on the full development set of SQuAD, i.e., our QA test + development set.}
\label{tab:qa-comp}
\vspace{-5pt}
\end{table}

\begin{table*}[t!]
\begin{center}
\small
\begin{tabular}{lllllll|l}
\toprule  & BLEU4 & METEOR & ROUGE-L & Q-BLEU1 & QPP & QAP & QA-Eval (H1) \\ 
\midrule
\citet{du2018harvesting} & 15.16 & 19.12 & -- & -- & -- & -- &  55.11/66.40\\
\midrule
Our baseline (w. BERT) & 18.05 & 22.41 & 46.57 & 49.38 & 29.08 & 54.61 & 58.63/69.97\\
 + QPP & 18.51 & 22.87 & 46.65 & 49.97 & \bf 30.14 & 55.67  &  \bf 60.49/71.81 \\
 + QAP & \bf 18.65 & \bf 22.91 & \bf 46.76 & \bf 50.09 & 30.09 & \bf 57.51 &  60.12/71.14\\
 + QPP \& QAP & 18.58 & 22.87 & 46.76 & 50.01 & 30.10 & 56.39 & 59.11/70.87 \\
\bottomrule
\end{tabular}
\end{center}
\vspace{-12pt}
\caption{The performance of our stronger BERT-QG models.} 
\vspace{-7pt}
\label{tab-bert-qg}
\end{table*}

\begin{table}[t!]
\begin{center}
\small
\begin{tabular}{lllll}
\toprule & Data & Data Size & EM & F1  \\ 
\midrule
\multirow{3}{*}{\rotatebox{90} {Dev set}} & SQuAD & 87k & 81.88 & 88.80  \\
& + Beam10  & 668k & 82.34 & 88.97 \\
& + H10 & 664k  & \bf 82.88 & \bf 89.53 \\
\midrule
\multirow{4}{*}{\rotatebox{90} {Test set}} & SQuAD & 87k & 80.25 & 88.23  \\
& + Beam10 & 668k & 81.20 & 88.36\\
& + H10 & 664k & 81.03 & \bf 88.79 \\
& + Beam10 + H10 & 1.2M & \bf 81.44  & 88.72 \\
\toprule
\end{tabular}
\end{center}
\vspace{-12pt}
\caption{The results of our semi-supervised QA method using a stronger BERT-QA model.}
\label{tab-bert-qa}
\vspace{-7pt}
\end{table}

\paragraph{Semi-Supervised QA results} Table~\ref{tab:semi-res} demonstrates the semi-supervised QA results. Without introducing new articles, we keep beam search outputs as additional questions. It can be seen that using beam search with beam size 10 (+Beam10) improves the BiDAF-QA baseline by 1.51/1.13 absolute points on the testing set. With introducing new articles, the best performance (+H8) improves the BiDAF-QA baseline by 1.69/1.27 absolute points on the testing set. We also combine the two best settings (Beam10+H8), but it does not perform better than using them separately.

We conduct two ablation studies on the development set. First, we compare beam search with different beam sizes and diverse beam search \cite{li2016simple}, but all of them perform similarly. Second, increasing the size of synthetic data from H1 to H10, the performance saturates around H2-H4. We also observed that when using a big synthetic dataset, e.g., H10, the model converges even before all examples were used for training. Based on these results, we conjecture that there is an upper bound of the effect of synthetic data which might be limited by the QG quality. To further improve the performance, more diverse and tricky questions need to be generated. To show how QG models help or limit the QA performance, we include some synthetic QA examples in Appendix. %
Finally, we compare our semi-supervised QA methods with \citet{dhingra2018simple}. As shown in Table~\ref{tab:qa-comp}, with no or less new data injection, our methods achieve larger improvements over a stronger baseline than their method. 

\subsection{QG and QA Results with BERT}

The Bidirectional Encoder Representations from Transformers (BERT) \cite{devlin2018bert} has recently improved a lot of NLP tasks by substantial margins. To verify if our improvements still hold on BERT-based baselines, we propose a BERT-QG baseline and test our two semantics-enhanced rewards; further, we conduct our semi-supervised QA method on a BERT-QA baseline.

\paragraph{BERT-QG} Without modifying our QG model's architecture, we simply replaced ELMo used above with BERT. Table~\ref{tab-bert-qg} shows that our BERT-QG baseline improves previous ELMo-QG baseline by a large margin; meanwhile, our QPP/QAP rewards significantly improve the stronger QG baseline and achieve the new state-of-the-art QG performance w.r.t both traditional metrics and QA-based evaluation. One difference is that the QAP-only model has the overall best performance instead of the multi-reward model.
Note that, we also obtain the QPP and QAP rewards from BERT-based QPC and QA models, respectively. 

\paragraph{BERT-QA} Using our QAP-reinforced BERT-QG model, we apply the same semi-supervised QA method on a BERT-QA baseline. As shown in Table~\ref{tab-bert-qa}, though with smaller margins, our method improves the strong BERT-QA baseline by 1.19/0.56 absolute points on testing set; even without introducing new articles, it obtains 0.95/0.13 absolute gains.

\section{Conclusion}
We proposed two semantics-enhanced rewards to regularize a QG model to generate semantically valid questions, and introduced a QA-based evaluation method that directly evaluates a QG model's ability to mimic human annotators in generating QA training data. Experiments showed that our QG model achieves new state-of-the-art performances. Further, we investigated how to use our QG system to augment QA datasets and conduct semi-supervised QA via two synthetic data generation methods along with a data filter and mixing mini-batch training. Experiments showed that our approach improves both BiDAF and BERT QA baselines even without introducing new articles.

\section*{Acknowledgments}
We thank the reviewers for their helpful comments and Hao Tan for useful discussions. This work was supported by DARPA (YFA17-D17AP00022), NSF-CAREER Award \#1846185, ONR Grant \#N00014-18-1-2871, and faculty awards from Google, Facebook, and Salesforce. The views contained in this article are those of the authors and not of the funding agency.

\bibliography{emnlp-ijcnlp-2019}
\bibliographystyle{acl_natbib}

\appendix
\section*{Appendix}
\section{Experiment Setup}
\subsection{Dataset}
\paragraph{QG} For QG, we use the SQuAD-based QG dataset\footnote{\url{https://github.com/xinyadu/nqg/tree/master/data}} first introduced by \citet{du2017learning} which was the most widely-used QG dataset in previous works \cite{song2018leveraging, zhao2018paragraph, du2018harvesting, kim2018improving, sun2018answer}. It was derived from SQuADv1.1 \cite{rajpurkar2016squad}. Since the testing set is not open, they sampled 10\% articles from the training set as the testing set, and the original development set is still used for validation.

For the QA-based QG evaluation, we obtain new paragraphs with pre-extracted answer spans from HarvestingQA \cite{du2018harvesting}. Without using their provided questions, we have different QG models act as ``annotators'' to generate questions, and then use the different QG-labeled synthetic datasets to train QA models. We use the same dev-test setup as described below.

\paragraph{QA} For QA, we use SQuADv1.1 \cite{rajpurkar2016squad}. Previous semi-supervised QA works sampled 10\% from training set as the testing set \cite{yang2017semi, dhingra2018simple}. Since we want to use the full training set in semi-supervised QA setup without any data size reduction, we instead split the original development set in half for validation and testing respectively. 

For semi-supervised QA, first, without introducing new articles, we generate new questions for SQuAD training set by keeping all beam search outputs. Second, with introducing new articles, we obtain new paragraphs with pre-extracted answer spans from HarvestingQA \cite{du2018harvesting}. Without using their provided questions, we use our best QG model to label questions. Meanwhile, we investigate the influence of synthetic data size, so we sample 10\% to 100\% examples from HarvestingQA, which are denoted as H1-H10 in our experiments.

\subsection{Evaluation Metrics}
\paragraph{QG} First, we use three traditional automatic evaluation metrics: BLEU4 \cite{papineni2002bleu}, METEOR \cite{denkowski2014meteor}, ROUGE-L \cite{lin2004rouge}. Second, we adopt the new ``Q-metrics'' proposed by \citet{nema2018towards}, and we only use ``Q-BLEU1'' that was shown to have the highest correlation with human judgments on SQuAD. We also take the QPP and QAP rewards as two additional evaluation metrics. Further, we conduct a pairwise human comparison between our baseline and best QG models. Detailed human evaluation setup is described in the next section. For the QA-based QG evaluation, we use the same QA evaluation metrics as follows.

\paragraph{QA} Following the standard evaluation method for SQuADv1.1 \cite{rajpurkar2016squad}, we use Exact Match (EM) and F1 as two metrics.  

\begin{figure*}[t!]
\begin{center}
\small
\begin{tabularx}{\textwidth}{X}
\toprule Examples generated on SQuAD   \\
\midrule 
Context: ...new york city consists of \underline{\textbf{five}} boroughs, each of which is a
 separate county of new york state...  \\
Ground-truth: how many boroughs does new york city contain ? \\
ELMo-QG: how many boroughs make up new york city ? \\
BERT-QG: new york city consists of how many boroughs ? \\
\midrule
Context: ...gendün gyatso traveled in exile looking for allies. however, it was not until \underline{\textbf{1518}} that the secular phagmodru ruler captured lhasa from the rinbung, and thereafter the gelug was given rights to conduct the new years prayer...  \\
Ground-truth: when was gelug was given the right to conduct the new years prayer ?
 \\
ELMo-QG: in what year did the secular phagmodru ruler take over lhasa ? \\
BERT-QG: when did the secular phagmodru ruler capture lhasa from the rinbung ? \\
\midrule
Context: ...chopin attended the lower rhenish music festival in aix-la-chapelle with hiller, and it was there that chopin met felix mendelssohn. after the festival, the three visited düsseldorf... they spent what mendelssohn described as ``a very agreeable day'', \underline{\textbf{playing and discussing music}} at his piano...   \\
Ground-truth: what two activities did frédéric do while visiting for a day in düsseldorf with mendelssohn and hiller ?
 \\
ELMo-QG: what did mendelssohn do at his piano ? \\
BERT-QG: what did chopin do at his piano ? \\
\midrule
Context: ...to limit protests, officials pushed parents to sign a document, which forbade them from holding protests, in exchange of money, but some who refused to sign \underline{\textbf{were threatened}}...   \\
Ground-truth: what has happened to some who refuse to agree to not protest ?
 \\
ELMo-QG: what did some who refused to sign ? \\
BERT-QG: what did the officials refused to sign ? \\
\toprule
\toprule Examples generated on HarvestingQA \\
\midrule 
Context: ...nigeria prior to independence was faced with sectarian tensions and violence... some of the ethnic groups like the ogoni, have experienced severe environmental degradation due to \underline{\textbf{petroleum extraction}}...  \\
\citet{du2018harvesting}: what is the main reason for the ethnic groups ?
 \\
ELMo-QG: why has nigeria experienced severe environmental degradation ? \\
BERT-QG: why have the ogoni experienced severe environmental degradation ?
 \\
\midrule
Context: ...vietnam is located on the eastern indochina peninsula... at its narrowest point in the central \underline{\textbf{quảng bình province}}, the country is as little as across...  \\
\citet{du2018harvesting}: where is the country 's country located ?
 \\
ELMo-QG: in what province is vietnam located ? \\
BERT-QG: what province is vietnam 's narrowest point ? \\
\midrule
Context: ...the ottoman islamic legal system was set up differently from \underline{\textbf{traditional european courts}}...  \\
\citet{du2018harvesting}: where was the ottoman islamic legal system set ?\\
ELMo-QG: the ottoman islamic legal system was set up from what ? \\
BERT-QG: what was the ottoman islamic legal system set up differently from ? \\
\midrule
Context: ...the eastern shore of virginia is the site of \underline{\textbf{wallops flight facility}}, a rocket testing center owned by nasa...  \\
\citet{du2018harvesting}: what is the eastern shore of virginia owned by ? \\
ELMo-QG: what facility is owned by nasa ?\\
BERT-QG: what is the name of the rocket facility located by nasa ? \\
\toprule
\end{tabularx}
\end{center}
\vspace{-10pt}
\caption{Some synthetic QA examples generated by our QG models. \vspace{-10pt}}
\label{tab-example1}
\end{figure*}

\subsection{Human Evaluation}
We performed pairwise human evaluation between our baseline and the QPP\&QAP multi-reward model on Amazon Mechanical Turk. We
selected human annotators that are located in the US, have an approval rate greater than 98\%, and have at least 10,000 approved HITs. We showed the annotators an input paragraph with the answer bold in the paragraph and two questions generated by two QG models (randomly shuffled to anonymize model identities). We then asked them to decide which one is better or choose ``non-distinguishable'' if they are equally good/bad. We give human three instructions about what is a good question: first, ``answerability'' -- a good question should be answerable by the given answer; ``making sense'' -- a good question should be making sense given the surrounding context; ``overall quality'' -- a good question should be as fluent, non-ambiguous, semantically compact as possible. Ground-truth questions were not provided to avoid simple matching with ground-truth.

\section{Implementation Details}
\paragraph{QG} For ELMo-QG, we first tokenize and obtain the POS/NER tags by Standford Corenlp toolkit\footnote{\url{https://stanfordnlp.github.io/CoreNLP/}}, then lower-case the entire dataset. We use 2-layer LSTM-RNNs for both encoder and decoder with hidden size 600. Dropout with a probability of 0.3 is applied to the input of each LSTM-RNN layer. We use the pre-trained character-level word embedding from ELMo \cite{peters2018deep} both as our word embedding and output-projection matrix, and keep it fixed. We use Adam \cite{kingma2014adam} as optimizer with learning rate 0.001 for teacher forcing and 0.00001 for reinforcement learning. Batch size is set to 32. For stability, we first pre-train the model with teacher forcing until convergence, then fine-tune it with the mixed loss. Hyper-parameters are tuned on development set: $\gamma^{qpp}=0.99$, $\gamma^{qap}=0.97$, and $n:m=3:1$. We use beam search with beam size 10 for decoding and apply a bi-gram/tri-gram repetition penalty as proposed in \citet{paulus2017deep}. 

For BERT-QG, we simply replace the ELMo used above to BERT \cite{devlin2018bert}. To match with BERT's tokenization, we use the WordPiece tokenizer to tokenize each word obtained above and extend the POS/NER tags to each word piece. Decoder's word-piece outputs will be mapped to normal words by post-processing. Hyper-parameters are tuned on development set: $\gamma^{qpp}=0.99$, $\gamma^{qap}=0.97$, and $n:m=1:3$. 

\paragraph{QA} For BiDAF-QA, we implement the BiDAF+Self-attention architecture proposed by \citet{clark2018simple}. We use GRUs for all RNN layers with hidden size 90 for GRUs and 180 for linear layers. Dropout with a probability of 0.2 is applied to the input of each GRU-RNN layer. We optimize the model using Adadelta with batch size 64. We also add ELMo to both the input and output of the contextual GRU-RNN layer as proposed in \cite{peters2018deep}. To match with QG model's setup, we also apply lower-case on QA datasets. 

For BERT-QA, we use the pre-trained uncased BERT-base model\footnote{\url{https://github.com/google-research/bert}} and fine-tune it on QA datasets. 

\paragraph{QPC}
For ELMo-QPC, we follow the model architecture proposed by \citet{conneau2017supervised}. First, two input questions are embedded with ELMo \cite{peters2018deep}. Second, the embedded questions are encoded by two 2-layer bidirectional LSTM-RNNs separately with hidden size 512. Next, a max-pooling layer outputs the sentence embedding of each question, denoted by $q_1$ and $q_2$. Lastly, we input $[q_1, q_2, |q_1 - q_2|, q_1*q_2]$ to an MLP to predict whether these two questions are paraphrases or not. This QPC model is trained using the Quora Question Pairs\footnote{\url{https://tinyurl.com/y2y8u5ed}} dataset. We use Adam \cite{kingma2014adam} as optimizer with learning rate 0.0004 and batch size 64. This model obtained 86\% accuracy on QQP development set. 

For BERT-QPC, we also use the pre-trained uncased BERT-base model and fine-tune it on QQP dataset, which obtained 90\% accuracy on QQP development set.

\section{Examples}
Figure~\ref{tab-example1} shows some synthetic QA examples generated by our QG models. On SQuAD, the first two examples show our QG models generate some paraphrases or novel questions that enrich the dataset; the last two examples show our QG models generate easier or wrong questions that limit the semi-supervised QA's performance. On HarvestingQA, our QG models can output better questions than \citet{du2018harvesting} did but still generate some wrong questions.

\end{document}